\documentclass[11pt]{article}

\usepackage{epsfig}
\usepackage{subfigure}
\usepackage{wrapfig}
\usepackage{url}
\usepackage{apa-uiuc}
\usepackage{enumerate}
\usepackage{multirow}
\usepackage{color}
\usepackage{amsmath}

\definecolor{myblue}{rgb}{0.165,0.34,0.5}

\newcommand{\titlestring}{Distance-Based Bias in Model-Directed Optimization of Additively Decomposable Problems}
\newcommand{\titlestringtwo}{Distance-Based Bias in Model-Directed Optimization of Additively Decomposable\\Problems}
\newcommand{\shortauthors}{Martin Pelikan and Mark W. Hauschild}
\newcommand{\reportnumber}{2012001}
\newcommand{\datestring}{January 2012}

\newcommand{\mysubsection}[1]{\subsection{#1}}

\begin{document}

\begin{sloppy}

\begin{titlepage}
\setlength{\parindent}{0pt}

\noindent
\includegraphics[width=5in]{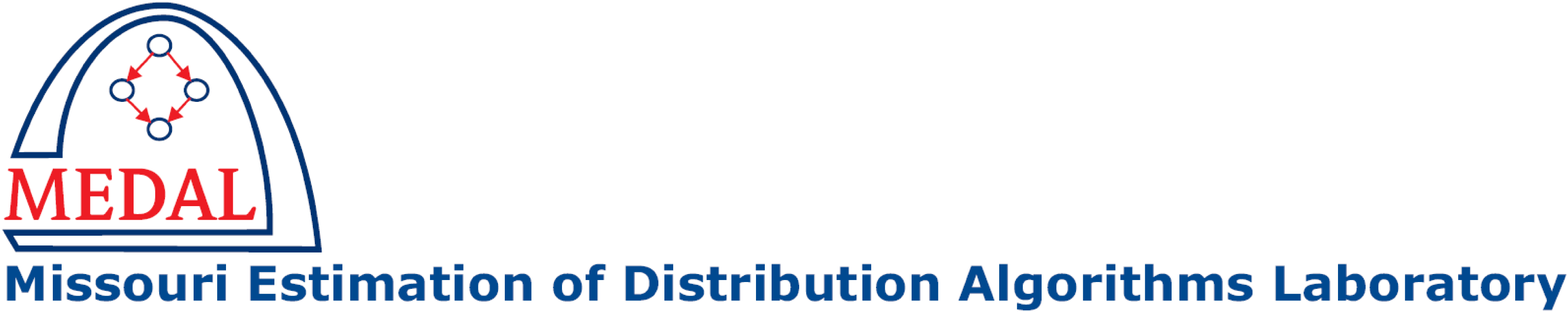}
\vspace*{0.075in}
{\color{myblue}
\hrule height 2pt
}
\vspace*{0.5in}

{\bf
\textsf{{\large
\titlestringtwo}}
}

\vspace*{0.25in}

\textsf{\shortauthors}

\vspace*{0.25in}

\textsf{MEDAL Report No. \reportnumber}

\vspace*{0.25in}

\textsf{\datestring}

\vspace*{0.25in}

{\bf \textsf{Abstract}}  

\vspace*{0.075in}

{\small \textsf{For many optimization problems it is possible to define a distance metric between problem variables that correlates with the likelihood and strength of interactions between the variables. For example, one may define a metric so that the dependencies between variables that are closer to each other with respect to the metric are expected to be stronger than the dependencies between variables that are further apart. The purpose of this paper is to describe a method that combines such a problem-specific distance metric with information mined from probabilistic models obtained in previous runs of estimation of distribution algorithms with the goal of solving future problem instances of similar type with increased speed, accuracy and reliability. While the focus of the paper is on additively decomposable problems and the hierarchical Bayesian optimization algorithm, it should be straightforward to generalize the approach to other model-directed optimization techniques and other problem classes. Compared to other techniques for learning from experience put forward in the past, the proposed technique is both more practical and more broadly applicable.}}

\vspace*{0.25in}

{\bf \textsf{Keywords}}

\vspace*{0.075in}
{\small \textsf{Estimation of distribution algorithms, hierarchical Bayesian optimization algorithm, decomposable problems, inductive transfer, learning from experience, efficiency enhancement.}}

\vfill

\noindent
\begin{minipage}{6in}
{\small \textsf{Missouri Estimation of Distribution Algorithms Laboratory (MEDAL)\\
Department of Mathematics and Computer Science\\
University of Missouri--St. Louis\\
One University Blvd.,
St. Louis, MO 63121\\
E-mail: \url{medal@cs.umsl.edu}\\
WWW: \url{http://medal.cs.umsl.edu/}\\}}
\end{minipage}

\end{titlepage}

\title{\titlestring}

\author{
{\bf Martin Pelikan}\\
Missouri Estimation of Distribution Algorithms Laboratory (MEDAL)\\
Dept. of Mathematics and Computer Science, 320 ESH\\
University of Missouri in St. Louis\\
One University Blvd., St. Louis, MO 63121\\
\url{pelikan@cs.umsl.edu}
\and
{\bf Mark W. Hauschild}\\
Missouri Estimation of Distribution Algorithms Laboratory (MEDAL)\\
Dept. of Mathematics and Computer Science, 321 ESH\\
University of Missouri in St. Louis\\
One University Blvd., St. Louis, MO 63121\\
\url{mwh308@umsl.edu}
}


\maketitle


\begin{abstract}
For many optimization problems it is possible to define a distance metric between problem variables that correlates with the likelihood and strength of interactions between the variables. For example, one may define a metric so that the dependencies between variables that are closer to each other with respect to the metric are expected to be stronger than the dependencies between variables that are further apart. The purpose of this paper is to describe a method that combines such a problem-specific distance metric with information mined from probabilistic models obtained in previous runs of estimation of distribution algorithms with the goal of solving future problem instances of similar type with increased speed, accuracy and reliability. While the focus of the paper is on additively decomposable problems and the hierarchical Bayesian optimization algorithm, it should be straightforward to generalize the approach to other model-directed optimization techniques and other problem classes. Compared to other techniques for learning from experience put forward in the past, the proposed technique is both more practical and more broadly applicable.
\end{abstract}


\section{Introduction}

Even for optimization problems that are extremely difficult to solve, it may be straightforward to extract information about important dependencies between variables and other problem regularities directly from the problem definition~\cite{Baluja:06,Drezner:02,Hauschild:10,Stonedahl:08,Schwarz:00*}. Furthermore, when solving many problem instances of similar type, it may be possible to gather information about variable interactions and other problem features by examining previous runs of the optimization algorithm, and to use this information to bias optimization of future problem instances to increase its speed, accuracy and reliability~\cite{Hauschild:08,Hauschild:11b}. 
The use of information from previous runs to introduce bias into future runs of an evolutionary algorithm is often referred to as {\em learning from experience}~\cite{Hauschild:08,Hauschild:11b,Pelikan:thesis}. 
The use of bias based on the results of other learning tasks in the same problem domain is also commonplace in machine learning where it is referred to as {\em inductive transfer} or {\em transfer learning}~\cite{Pratt:91,Caruana:97}. Numerous studies have shown that using prior knowledge and learning from experience promise improved efficiency and problem solving capabilities~\cite{Baluja:06,Drezner:02,Hauschild:10,Hauschild:11b,Rothlauf:06,Stonedahl:08,Schwarz:00*}. However, most prior work in this area was based on hand-crafted search operators, model restrictions, or representations.  

This paper describes an approach that combines prior problem-specific knowledge with learning from experience. The basic idea of the proposed approach comprises of (1)~defining a problem-specific distance metric, (2)~analyzing previous EDA models to quantify the likelihood and nature of dependencies at various distances, and (3)~introducing bias into EDA model building based on the results of the analysis using Bayesian statistics. One of the key goals of this paper is to develop an automated procedure capable of introducing bias based on a distance metric and prior EDA runs, without requiring much expert knowledge or hand-crafted model restrictions from the practitioner. Furthermore, the proposed approach is intended to be applicable in a more practical manner than other approaches to learning from experience. For example, the proposed approach makes it feasible to use prior runs on problems of a smaller size to introduce bias when solving problem instances of a larger size, and the bias can be introduced even when the importance of dependencies between specific pairs of variables varies significantly across the problem domain. 
Although this paper focuses on the hierarchical Bayesian optimization algorithm (hBOA) and additively decomposable functions (ADFs), the proposed approach can also be applied to other model-directed optimization techniques and other problem types. The paper outlines a framework that can be used to adapt the proposed approach to a different context.

The paper is organized as follows. Section~\ref{section-hboa} describes hBOA. Section~\ref{section-framework} outlines the framework for introducing bias based on prior runs on similar problems in model-directed optimization. Section~\ref{section-bias} presents the proposed approach to introducing bias into hBOA model building for additively decomposable problems. Section~\ref{section-experiments} presents experimental results. Section~\ref{section-conclusions} summarizes and concludes the paper.


\section{Hierarchical BOA}
\label{section-hboa}


The hierarchical Bayesian optimization algorithm (hBOA)~\cite{Pelikan:01*,Pelikan:03b,Pelikan:book} is an estimation of distribution algorithm (EDA)~\shortcite{Baluja:94,Larranaga:02,Pelikan:02,Larranaga:06,Pelikan:EDA-book,Hauschild:11c}. hBOA works with a population of candidate solutions represented by fixed-length strings over a finite alphabet. In this paper, candidate solutions are represented by $n$-bit binary strings. The initial population of binary strings is generated at random according to the uniform distribution over candidate solutions. Each iteration starts by selecting promising solutions from the current population; here binary tournament selection without replacement is used. Next, hBOA (1)~learns a Bayesian network with local structures for the selected solutions and (2)~generates new candidate solutions by sampling the distribution encoded by the built network~\cite{Chickering:97,Friedman:99}. To maintain useful diversity in the population, the new candidate solutions are incorporated into the original population using restricted tournament selection (RTS)~\cite{Harik:95a}. The run is terminated when termination criteria are met. In this paper, each run is terminated either when the global optimum is found or when a maximum number of iterations is reached. 
Since the basic understanding of probabilistic models used in hBOA is necessary for the remainder of the paper, the rest of this section discusses the class of probabilistic models used in hBOA. 

hBOA represents probabilistic models of candidate solutions by Bayesian networks with local structures~\cite{Chickering:97,Friedman:99}. A Bayesian network is defined by two components: (1)~an acyclic directed graph over problem variables specifying direct dependencies between variables and (2)~conditional probabilities specifying the probability distribution of each variable given the values of the variable's parents. A Bayesian network encodes a joint probability distribution as $p(X_1,\ldots,X_n)=\prod_{i=1}^n p(X_i|\Pi_i)$ where $X_i$ is the $i$th variable and $\Pi_i$ are the parents of $X_i$ in the underlying graph. 

To represent conditional probabilities of each variable given the variable's parents, hBOA uses decision trees. Each internal node of a decision tree specifies a variable, and the subtrees of the node correspond to the different values of the variable. Each leaf of the decision tree for a particular variable defines the probability distribution of the variable given a condition specified by the constraints given by the path from the root of the tree to this leaf (constraints are given by the assignments of the variables along this path).  

To build probabilistic models, hBOA typically uses a greedy algorithm that initializes the decision tree for each problem variable $X_i$ to a single-node tree that encodes the unconditional probability distribution of $X_i$. In each iteration, the model building algorithm tests how much a model would improve after splitting each leaf of each decision tree on each variable that is not already located on the path to the leaf. The algorithm executes the split that provides the most improvement, and the process is repeated until no more improvement is possible.

Improvement of the model after a split is often evaluated using the Bayesian-Dirichlet (BDe) metric with penalty for model complexity. Bayesian measures evaluate the goodness of a Bayesian network structure given data $D$ and background knowledge $\xi$ as~\cite{Cooper:92,Heckerman+al:94}
\begin{equation}
\label{eq-metric}
p(B|D,\xi) = c p(B|\xi) p(D|B,\xi),
\end{equation}

\noindent
where $c$ is a normalization constant. For the Bayesian-Dirichlet metric, the term $p(D|B,\xi)$ is estimated as~\cite{Chickering:97}
\begin{equation}
\label{eq-metric2}
\begin{split}
p(D|B,\xi)=
\prod_{i=1}^{n}
\prod_{l\in L_i}
\frac{\Gamma(m_i'(l))}{\Gamma(m_i(l)+m_i'(l))}\\
\prod_{x_i}
\frac{\Gamma(m_i(x_i,l)+m_i'(x_i,l))}{\Gamma(m_i'(x_i,l))},
\end{split}
\end{equation}

\noindent
where $L_i$ is the set of leaves in the decision tree $T_i$ for $X_i$; $m_i(l)$ is the number of instances in the selected population which end up the traversal through the tree $T_i$ in the leaf $l$; $m_i(x_i,l)$ is the number of instances that have $X_i=x_i$ and end up the traversal of the tree $T_i$ in the leaf $l$; $m_i'(l)$ represents the prior knowledge about the value of $m_i(i,l)$; and $m_i'(x_i,l)$ represents the prior knowledge about the value of $m_i(x_i,l)$. Without any prior knowledge, an uninformative prior $m'_i(x_i,l)=1$ is typically used. 
To favor simpler networks to the more complex ones, the prior probability of each network decreases exponentially fast with respect to the description length of this network's parameters~\cite{Friedman:99,Pelikan:book}:
\begin{equation}
p(B) = c 2^{-0.5 (\sum_{i} |L_i|) \log_2 N},
\end{equation}

\noindent
where $c$ is a normalization constant required for the prior probabilities of all network structures to sum to one.




\section{Bias Based on Previous EDA Runs}

\label{section-framework}
Building an accurate probabilistic model in hBOA and other EDAs  based on complex probabilistic models can be time consuming and it may require rather large populations of solutions. That is why much effort has been put into enhancing efficiency of model building in EDAs and improving quality of EDA models even with smaller populations of candidate solutions~\cite{Baluja:06,Hauschild:08,Hauschild:09b,Hauschild:11b,Muhlenbein:02}. 
 Learning from experience~\cite{Hauschild:08,Hauschild:09b,Hauschild:11b,Pelikan:book} represents one approach to dealing with this issue. In learning from experience, models discovered by EDAs in previous runs are mined to identify regularities and the discovered regularities are used to bias model building in future runs on problems of similar type. Since learning model structure is often the most challenging task in model building, learning from experience often focuses on identifying regularities in model structure and using these regularities to bias structural learning in future runs. 

It is straightforward to collect statistics on the most frequent dependencies in EDA models. Nonetheless, for the collected statistics to be useful, it is important to ensure that the statistics are meaningful with respect to the problem being solved. For example, consider 
optimization of NK landscapes~\cite{Kauffman:89}, in which the fitness function is defined as the sum of $n$ subfunctions $\{f_i\}^n_{i=1}$, and the subfunction $f_i$ is applied to the $i$th bit and its $k$ neighbors. The neighbors of each bit are typically chosen at random for each problem instance. Therefore, if we consider 1,000 problem instances of NK landscapes, looking at the percentage of models that included a dependency between the first two bits for the first 999 instances will not say much about the likelihood of the same dependency for the last instance. A similar observation can be made for many other important problem classes, such as MAXSAT or the quadratic assignment problem. That is why it is important to develop a more general framework that allows one to learn and use statistics on dependencies in EDA models across a range of problem domains of different structure and properties. In the remainder of this section we describe one such framework. 

To formalize the proposed framework to identifying structural regularities in EDA models, let us define a set of $m$ dependency categories $D=\{D_1,\ldots,D_m\}$ and denote the background knowledge about the problem by $\xi$. Then, we can define a function $\gamma(i,j,\xi)$ that, given $\xi$, maps any dependency $(i,j)$ covered by the probabilistic model into one of the $m$ categories so that $\gamma(i,j,\xi) = k \mbox{~~if and only if~~} (i,j)\in D_k.$
Two straightforward possibilities for defining $\gamma$ function were proposed by Hauschild et al.~\citeyear{Hauschild:08,Hauschild:09b}: (1)~Each pair of problem variables $X_i$ and $X_j$ defines a special category, and (2)~categories are defined using a discretization of a problem-specific distance metric between variables. The first approach is useful especially when solving a number of instances of a problem where each variable has a fixed meaning across the entire set of instances; this is the case for example in spin glasses defined on a regular lattice, where every pair of variables in the lattice can be assigned a special category because the structure of the problem does not change from one instance to another~\cite{Hauschild:09c}. 
The latter approach is useful especially when one can define a distance metric on variables so that the distance between two variables correlates strongly with the likelihood or strength of their interaction; for example, one may define a distance metric such that variables that interact more strongly are closer to each other according to the metric. Such a distance metric can be defined for example in the quadratic assignment problem, traveling salesman problem or, more generally, classes of additively decomposable functions. 
While these two approaches are applicable to many important classes of problems, one may envision many other approaches based on this framework. The key issue in defining $\gamma$ is that the categories should be related to the problem, so that each category contains pairs of variables that have a lot in common and that can be expected to be either correlated or independent most of the time. 

The statistics obtained from previous EDA models can be used to bias the search operators of model-directed optimization methods using either a soft bias or a hard bias. A soft bias allows one to define preference to some models using a prior distribution over network structures or partial variable assignments~\cite{Schwarz:00*,Hauschild:09b}. A hard bias encodes hard restrictions on model structure or variable assignments, restricting the class of allowable models~\cite{Muhlenbein:02,Baluja:06,Hauschild:08}. While in most prior work on bias in EDAs the bias was based on expert knowledge, in learning from experience the focus is on {\em automated learning} of a bias from past EDA runs. 

In this paper we describe one way of using the above framework to facilitate learning from experience in hBOA for additively decomposable problems based on a problem-specific distance metric. However, note that the framework can be applied to other EDAs based on graphical models.



\section{Distance-Based Bias}
\label{section-bias}


\mysubsection{Additively Decomposable Functions}
For many optimization problems, the objective function (fitness function) can be expressed in the form of an additively decomposable function (ADF) of $m$ subproblems:

\begin{equation}
f(X_1,\ldots,X_n) = \sum_{i=1}^m f_i(S_i),
\end{equation}

\noindent
where $(X_1,\ldots,X_n)$ are problem's decision variables, $f_i$ is the $i$th subfunction, and $S_i\subset \{X_1,X_2,\ldots,X_n\}$ is the subset of variables contributing to $f_i$. While they may often exist multiple ways of decomposing the problem using additive decomposition, one would typically prefer decompositions that minimize the sizes of subsets $\{S_i\}$. It is of note that the difficulty of ADFs is not fully determined by the order of subproblems, but also by the definition of the subproblems and their interaction. In fact, there exist a number of NP-complete problems that can be formulated as ADFs with subproblems of order $2$ or $3$, such as MAXSAT for 3CNF formulas. On the other hand, one may easily define ADFs with subproblems of order $n$ that can be solved by a simple bit-flip hill climbing in low-order polynomial time. 

\mysubsection{Measuring Variable Distances for ADFs}
The definition of a distance between two variables of an ADF used in this paper follows~\citeN{Hauschild:08} and \shortciteN{Hauschild:11b}. Given an additively decomposable problem, we define the distance between two variables using a graph $G$ of $n$ nodes, one node per variable. For any two variables $X_i$ and $X_j$ in the same subset $S_k$, that is, $X_i, X_j\in S_k$, we create an edge in $G$ between the nodes $X_i$ and $X_j$. Denoting by $l_{i,j}$ the number of edges along the shortest path between $X_i$ and $X_j$ in $G$ (in terms of the number of edges), we define the distance between two variables as
\[
D(X_i,X_j) = 
\left\{
\begin{array}{ll}
l_{i,j} & \mbox{if a path between $X_i$ and $X_j$ exists}\\
n & \mbox{otherwise}
\end{array}
\right.
\]

\noindent
The above distance measure makes variables in the same subproblem close to each other, whereas for the remaining variables, the distances correspond to the length of the chain of subproblems that relate the two variables. The distance is maximal for variables that are completely independent (the value of a variable does not influence the contribution of the other variable in any way). 

Since interactions between problem variables are encoded mainly in the subproblems of the additive problem decomposition, the above distance metric should typically correspond closely to the likelihood of dependencies between problem variables in probabilistic models discovered by EDAs. Specifically, the variables located closer with respect to the metric should more likely interact with each other. Figure~\ref{fig-split-prob-at-dist} illustrates this on two ADFs discussed later in this paper---the NK landscape with nearest neighbor interactions and the two-dimensional Ising spin glass (for a description of these problems, see section~\ref{section-problems}). The figure analyzes probabilistic models discovered by hBOA in 10 independent runs on each of the 1,000 random instances for each problem and problem size.
For a range of distances $d$ between problem variables, the figure shows the proportion of splits on a variable located at distance $d$. The results clearly support the fact that hBOA models indicate strongest dependencies between variables located close to each other according to the aforementioned metric and that there is a clear correlation between the distance metric and the likelihood of dependencies. Furthermore, the figure indicates that the likelihood of dependencies at a specific distance does not change much from one problem size to another, indicating that the bias based on these statistics should be applicable across a range of problem sizes.

\begin{figure}
\centering
\subfigure[Nearest-neighbor NK.]{\epsfig{file=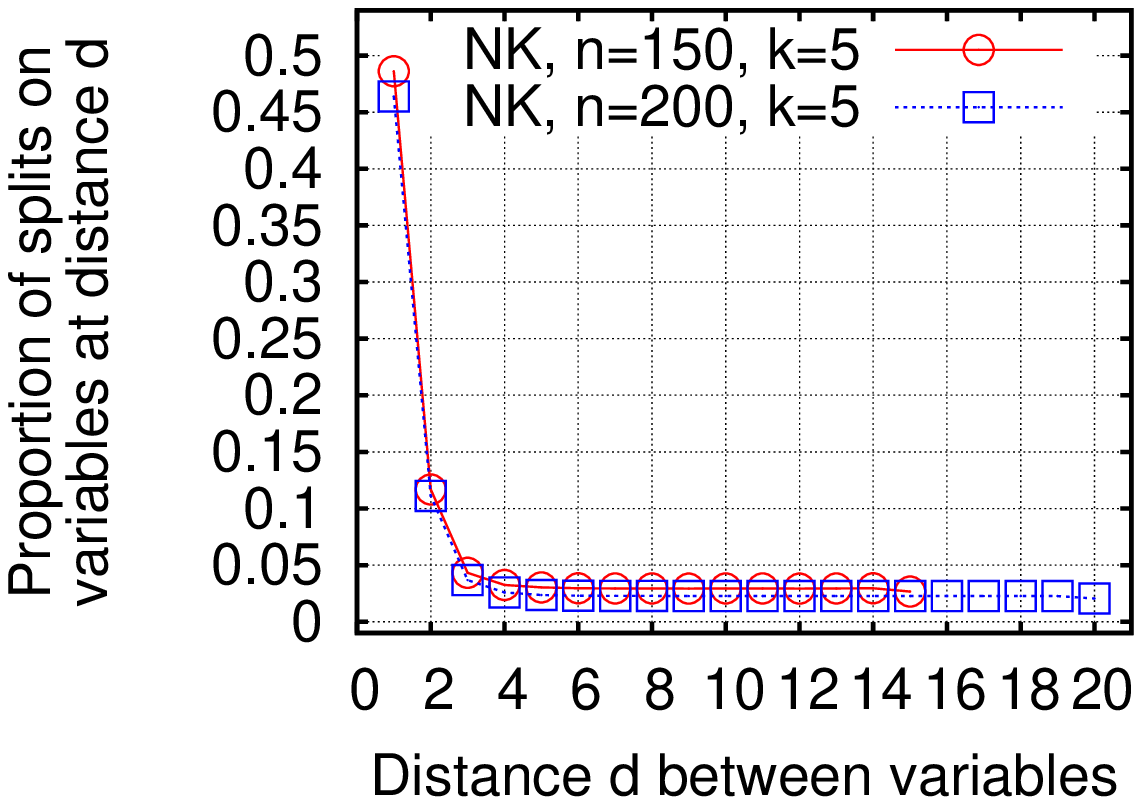,width=0.35\textwidth}}
\subfigure[2D spin glass.]{\epsfig{file=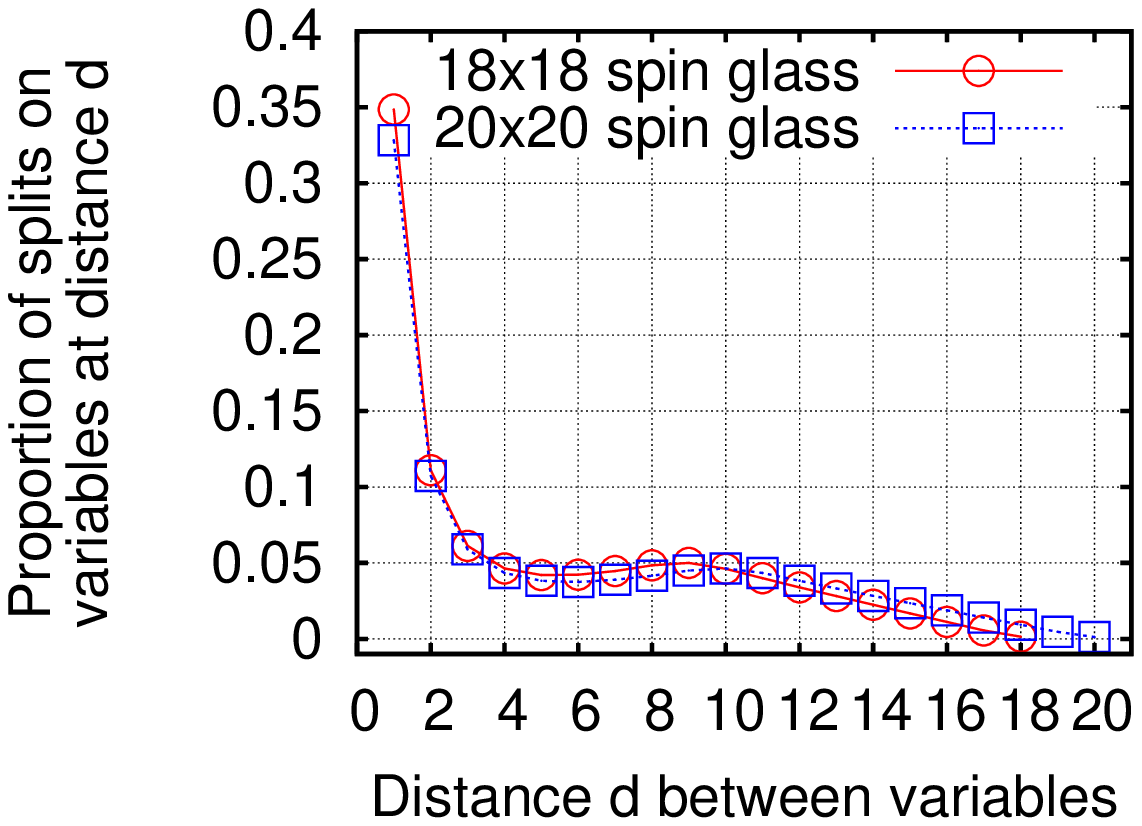,width=0.35\textwidth}}
\caption{\small Dependencies between variables that are closer to each other are stronger than the dependencies between other variables. Furthermore, the proportion of splits capturing dependencies between variables at a given distance changes only little between instances of various sizes. This indicates that the statistics acquired from models of one size should be useful for biasing hBOA runs on problems of another size.
The results are based on 979,020 models obtained from hBOA on 1,000 unique problem instances of each problem, 10 runs per instance.
}
\label{fig-split-prob-at-dist}
\end{figure}

It is important to note that other approaches may be envisioned to defining a distance metric for ADFs. For example, a weight may be added on each edge that would decrease with the number of subsets that contain the two connected variables. Another interesting possibility would be to consider the subfunctions themselves in measuring the distances, so that only correlations that lead to nonlinearities are considered or that some correlations are given a priority over others. Finally, the distance of variables may depend on the problem definition itself, not on the decomposition only. For example, in the quadratic assignment problem, a distance between two facility locations is specified directly by the problem instance. The key is to use problem-specific information to specify a distance metric so that the distance between a pair of variables correlates with the likelihood or strength of their interaction.


\mysubsection{Using Distance-Based Bias in hBOA}
The basic idea of incorporating the distance-based bias based on prior runs into hBOA is inspired mainly by the work of Hauschild et al.~\cite{Hauschild:09b}. Hauschild et al. proposed to incorporate learning from experience into hBOA by modifying prior probabilities of network structures 
using the statistics that capture the number of splits on each variable in the decision tree for each other variable in past hBOA runs on similar problems. 
Nonetheless, the approach of Hauschild et al.~\cite{Hauschild:09b} is only applicable to problems where the strength of interactions between any two variables is not expected to change much from instance to instance. That is why this approach can only be applied in a limited set of problem domains and it is difficult to use this approach when problem size is not fixed in all runs. In this paper, we propose to capture the {\em nature of dependencies between variables with respect to their distance} using the distance metric defined for ADFs or another distance metric. This allows one to apply the technique in more problem domains and also allows models from problem instances of one size to be useful in solving problem instances of different sizes (see figure~\ref{fig-split-prob-at-dist}).

Recall that the BDe metric used to evaluate probabilistic models in hBOA contains two parts: (1) prior probability $p(B|\xi)$ of the network structure $B$, and (2) posterior probability $p(D|B,\xi)$ of the data (population of selected solutions) given $B$. Prior probabilities of network structures are typically set to represent the uniform distribution over admissible network structures or to provide a bias toward simple models regardless of the problem. However, the prior probability distribution of network structures can also be used to specify preferable structures. 
In this paper, we will use the prior probability distribution of network structures to introduce bias toward models that resemble models obtained in previous runs on problems of similar type with the focus on distance-based statistics. An analogous approach can be used to incorporate bias into hBOA for a different mapping $\gamma$ of pairs of variables into dependency categories $\{D_i\}$.

Let us assume a set $M$ of hBOA models from prior hBOA runs on similar ADFs. Before applying the bias by modifying the prior probability distribution of models in hBOA, the models in $M$ are first processed to generate data that will serve as the basis for introducing the bias. The processing starts by analyzing the models in $M$ to determine the number $s(m,d,j)$ of splits on any variable $X_i$ such that $D(X_i,X_j)=d$ in a decision tree $T_j$ for variable $X_j$ for a model $m\in M$. Then, the values $s(m,d,j)$ are used to compute the probability $P_k(d,j)$ of a $k$th split on a variable at distance $d$ from $X_j$ in a dependency tree $T_j$ given that $k-1$ such splits were already performed in $T_j$:
\begin{equation}
P_k(d,j) = \frac{\left|\{m\in M: s(m,d,j)\geq k\}\right|}{\left|\{m\in M: s(m,d,j)\geq k-1\}\right|}\cdot
\end{equation}

\noindent
Given the terms $P_k(d,j)$, we can now define the prior probability of a network $B$ as
\begin{equation}
\label{eq-pb-bias}
p(B,\xi) = c \prod_{d=1}^n \prod_{j=1}^n \prod_{k=1}^{n_s(d,j)} P^{\kappa}_k(d,j),
\end{equation}

\noindent 
where $n_s(d,j)$ denotes the number of splits on any variable $X_i$ such that $D(X_i,X_j)=d$ in $T_j$, $\kappa>0$ is used to tune the strength of bias (the strength of bias increases with $\kappa$), and $c$ is a normalization constant. Since log-likelihood is typically used to evaluate model quality, to evaluate the contribution of any particular split, the main difference from the standard version of the BDe metric with the complexity penalty is that instead of reducing the metric according to the additional complexity of $\log_2(N)/2$ for each new split, we reduce the metric by the corresponding $\kappa \log_2 P_k(d,j)$. Therefore, the computation of the change of the prior probability of the network structure can still be done in constant time. Of course, the change in $p(D|B,\xi)$ requires computation of marginal frequencies, so it cannot be done in constant time. 

It is important to note that the prior probability of hBOA models defined in eq.~(\ref{eq-pb-bias}) is certainly not the only possible approach to incorporating learning from experience using distance-based statistics into hBOA. The main source of inspiration for the proposed approach is the work on incorporating bias in learning Bayesian networks using Bayesian metrics~\cite{Heckerman+al:94} and the prior work on learning from experience in hBOA by Hauschild et al.~\cite{Hauschild:09b}. The experimental results presented in the next section confirm that this approach leads to substantial speedups in both problem classes considered in this paper and preliminary experiments in other problem domains including MAXSAT and minimum vertex cover indicate that substantial speedups can be obtained also in other problem classes defined as ADFs. 


\section{Experiments}
\label{section-experiments}

\mysubsection{Test Problems}
\label{section-problems}
To test the proposed approach to biasing hBOA model building, we consider two problem classes: shuffled nearest-neighbor NK landscapes and two-dimensional spin glasses. Both these problem classes were shown to be challenging for conventional genetic algorithms and many other optimization techniques due to the rugged landscape, strong epistasis, and complex structure of interactions between problem variables~\cite{Kauffman:89,young1998,Pelikan:10,Pelikan:06}. However, for both problem classes, it is straightforward to generate a large number of problem instances with known optima. For each problem class and problem size, we use 1,000 unique problem instances; the reason for using such a large number of instances is that for these problem classes, algorithm performance often varies substantially from one instance to another and the results would thus be unreliable if only a few instances were used. 

An NK fitness landscape~\cite{Kauffman:89} is fully defined by the following components: (1) The number of bits, $n$, (2) the number of neighbors per bit, $k$, (3) a set of $k$ neighbors $\Pi(X_i)$ of the $i$th bit for every $i\in\{1,\ldots, n\}$, and (4) a subfunction $f_i$ defining a real value for each combination of values of $X_i$ and $\Pi(X_i)$ for every $i\in\{1,\ldots, n\}$. Typically, each subfunction is defined as a lookup table. 
The objective function $f_{nk}$ to maximize is defined as 
%
$f_{nk}(X_1,X_2,\ldots, X_{n}) = \sum_{i=1}^{n} f_i(X_i,\Pi(X_i)).$
%
The difficulty of optimizing NK landscapes depends on all components defining an NK problem instance. In this paper, we consider nearest-neighbor NK landscapes, in which neighbors of each bit are restricted to the $k$ bits that immediately follow this bit. The neighborhoods wrap around; thus, for bits which do not have $k$ bits to the right, the neighborhood is completed with the first few bits of solution strings. The reason for restricting neighborhoods to nearest neighbors was to ensure that the problem instances can be solved in polynomial time even for $k>1$ using dynamic programming~\cite{Pelikan:10}. The subfunctions are represented by look-up tables (a unique value is used for each instance of a bit and its neighbors), and each entry in the look-up table is generated with the uniform distribution from $[0,1)$. The used class of NK landscapes with nearest neighbors is thus the same as that in~\citeN{Pelikan:10}. In all experiments, we use $k=5$ and $n\in\{100,150,200\}$. For each $n$, we use 1,000 unique, independently generated instances; overall, 3,000 unique instances of NK landscapes were tested. 

Ising spin glasses are prototypical models for disordered systems~\cite{young1998}.
A simple model to describe a finite-dimensional
Ising spin glass is typically arranged on a
 regular 2D or 3D grid where each node $i$ corresponds to a spin $s_i$ 
and each edge $\langle i,j \rangle$
 corresponds to a coupling between two spins $s_i$ and $s_j$. Each edge has a
 real value $J_{i,j}$
 associated with it that defines the relationship between the two connected
 spins. To approximate the behavior of the large-scale system, periodic
 boundary conditions are often used that introduce a  coupling between the
 first and the last elements in each dimension.
For the classical Ising
model, each spin $s_i$ can be in one of two states: $s_i=+1$ or 
$s_i=-1$. Given a set of coupling constants
$J_{i,j}$, and a configuration of spins $C$, the energy can be
computed as
$E(C) = - \sum_{\langle i,j\rangle} s_i J_{i,j} s_j ,$
%
where the sum runs over all couplings $\langle i,j\rangle$. 
Here the task is to find a spin configuration for a given set of
coupling constants that minimizes the energy of the spin glass. The
states with minimum energy are called {\em ground states}. The spin
configurations are encoded with binary strings where each bit
specifies the value of one spin (0 for a spin +1, 1 for a spin -1). 
One generally analyzes a large set of random spin glass instances
for a given distribution of the spin-spin couplings. In this paper 
we consider the $\pm J$ spin glass, 
where each spin-spin coupling 
is set randomly to either $+1$ or $-1$ with equal probability. 
We use instances arranged on square grids of sizes $10\times 10$, $12\times 12$, $14\times 14$, $16\times 16$, $18\times 18$ and $20\times 20$ spins; that is,
the problem sizes range from $100$ to $400$ spins. We consider periodic boundary conditions. For each problem size,
we use 1,000 unique, independently generated problem instances; overall, 6,000 unique instances of the 2D spin glass were tested.
All instances were obtained from the
Spin Glass Ground State Server~\cite{SGS:Cologne}.

\mysubsection{10-Fold Crossvalidation}
To ensure that the same problem instances were not used for defining the bias as well as for testing it, 10-fold crossvalidation was used. 
For each problem size and each problem, 1,000 random problem instances were used in the experiments. The 1,000 instances in each set were randomly split into 10 equally sized subsets of 100 instances each. In each round of crossvalidation, 1 subset of 100 instances was left out and hBOA was run on the remaining 9 subsets of 900 instances total. The runs on the 9 subsets produced a number of models that were analyzed in order to obtain the probabilities $P_k(d,j)$ for all $d$, $j$, and $k$. The bias based on the obtained values of $P_k(d,j)$ was then used in hBOA runs on the remaining subset that was left out. The same procedure was repeated for each subset; overall, 10 rounds of crossvalidation were performed for each set of 1,000 instances. Each problem instance was used exactly once in the test of the proposed approach to biasing hBOA models and in every test, models used to generate statistics for hBOA bias were obtained from hBOA runs on different problem instances. 

While the experiments were performed across a variety of computer architectures and configurations, it was always ensured that the base case with no bias and the case with bias were both run on the same computational node and the results of the two runs can therefore be compared against each other with respect to the actual CPU time.

\mysubsection{Experimental Setup}
The maximum number of iterations for each problem instance was set to the overall number of bits in the problem; according to preliminary experiments, this upper bound substantially exceeded the actual number of iterations required to solve each problem. Each run was terminated either when the global optimum was found, when the population consisted of copies of a single candidate solution, or when the maximum number of iterations was reached. For each problem instance, we used bisection~\cite{Sastry:01c,Pelikan:book} to ensure that the population size was within $5\%$ of the minimum population size to find the optimum in 10 out of 10 independent runs. 

Bit-flip hill climbing (HC) is incorporated into hBOA to improve its performance. HC takes a candidate solution represented by an $n$-bit binary string on input. Then, it performs one-bit changes on the solution that lead to the maximum improvement of solution quality. HC is terminated when no single-bit flip improves solution quality and the solution is thus locally optimal. Here, HC is used to improve every solution in the population before the evaluation is performed. Without HC, the number and size of problem instances would have to substantially reduce due to the increased computational requirements. However, preliminary results indicate that even without HC the benefits of the proposed approach would be substantial. 
 
The proposed approach to distance-based bias in hBOA is tested for  $\kappa=\{1,2,3,4,5,6,7\}$ to assess how the strength of the bias represented by $\kappa$ affects hBOA performance.


To evaluate hBOA performance, we focus on (1) the execution time per run, (2) the number of steps of HC, (3) the number of evaluations, and (4) the required population size. The steps of HC are not counted as evaluations in order to distinguish between evaluations and HC steps, because for many additively decomposable problems, performing a HC step is much less computationally expensive than evaluating a solution. To evaluate the benefits of distance-based bias, the paper uses {\em multiplicative speedups}, where the speedup is defined as a multiplicative factor by which a particular complexity measure improves by using the distance-based bias compared to the base case with no distance-based bias. For example, an execution-time speedup of $2$ indicates that the bias allowed hBOA to find the optimum twice as fast as without the bias. Although the code could be further optimized for efficiency, the primary focus of our experiments  concerning the execution times was on the {\em speedups} of the CPU times rather than their absolute values. We have used the most efficient implementation of hBOA available for the base case with no bias and we only modified it for the remaining cases to incorporate the bias. 


\mysubsection{Results}
Fig.~\ref{fig-speedups1a} shows the effects of $\kappa$ on the multiplicative speedups with respect to the execution time, the number of evaluations, the number of HC steps, and the population size. The results confirm that, for adequate values of $\kappa$, the speedups in terms of execution time are substantial for both NK landscapes as well as 2D spin glasses; the maximum speedup for NK landscapes was over $2.26$ whereas for spin glasses it was over $1.66$. For NK landscapes, the speedups in terms of execution time grow both with problem size $n$ and with $\kappa$, and they can be expected to increase further for even larger values of $n$ or $\kappa$. For spin glasses, the speedups seem nearly independent of problem size and best speedups are obtained for $\kappa=3$. 
The multiplicative speedups in terms of the number of evaluations, the number of HC steps and the population size indicate that the reduction in the population sizes appears to be one of the most important factors reducing the overall computational cost, although improvements can be observed also in most other statistics.

Fig.~\ref{fig-speedups1b} shows the speedups obtained with respect to the problem size for a range of values of $\kappa$; these results are useful for visualizing the relationship between the problem size and the speedups obtained with the distance-based bias. The results confirm that for NK landscapes, the speedups appear to grow at least linearly with the problem size, regardless of the value of $\kappa$. However, for the 2D spin glass, the speedups fluctuate around the same value for all problem sizes. The speedups obtained on NK landscapes are thus not only better, but they further improve with problem size, unlike for the 2D spin glass. On one hand, one may argue that this is due to the fact that NK landscapes with nearest neighbor interactions have a simpler structure than the 2D spin glass due to the short-range interactions. On the other hand, the interactions in 2D spin glasses are of much smaller order (subproblems in problem decomposition have 2 bits each instead of 6). We are currently evaluating the distance-based bias on other classes of nearly decomposable problems such as MAXSAT and the minimum vertex cover in order to provide more empirical evidence that would help explain when the distance-based bias works better and when it has limitations. 

In summary, fig.~\ref{fig-speedups} provides solid empirical evidence that the speedups obtained are substantial and that the proposed approach to learning from experience is useful in practice. 

\begin{figure*}
\begin{center}
\epsfig{file=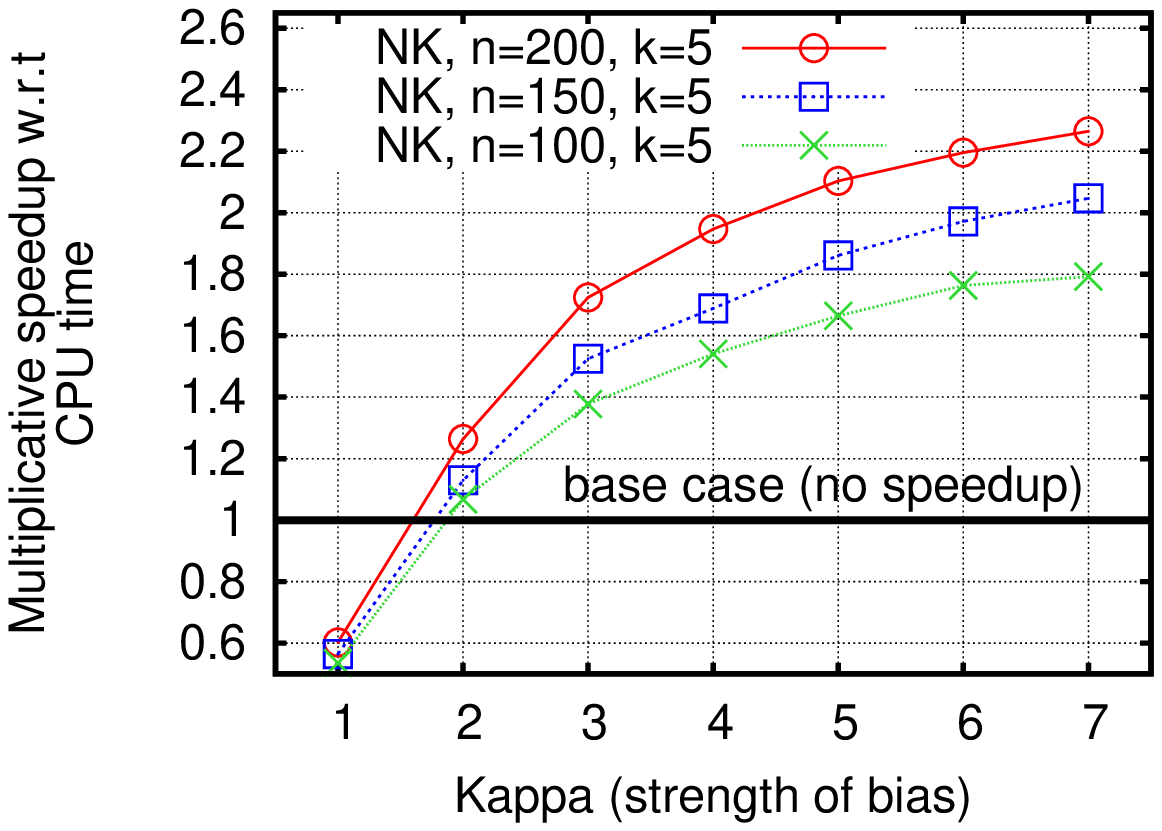,width=0.33\textwidth}\hspace*{5em}
\epsfig{file=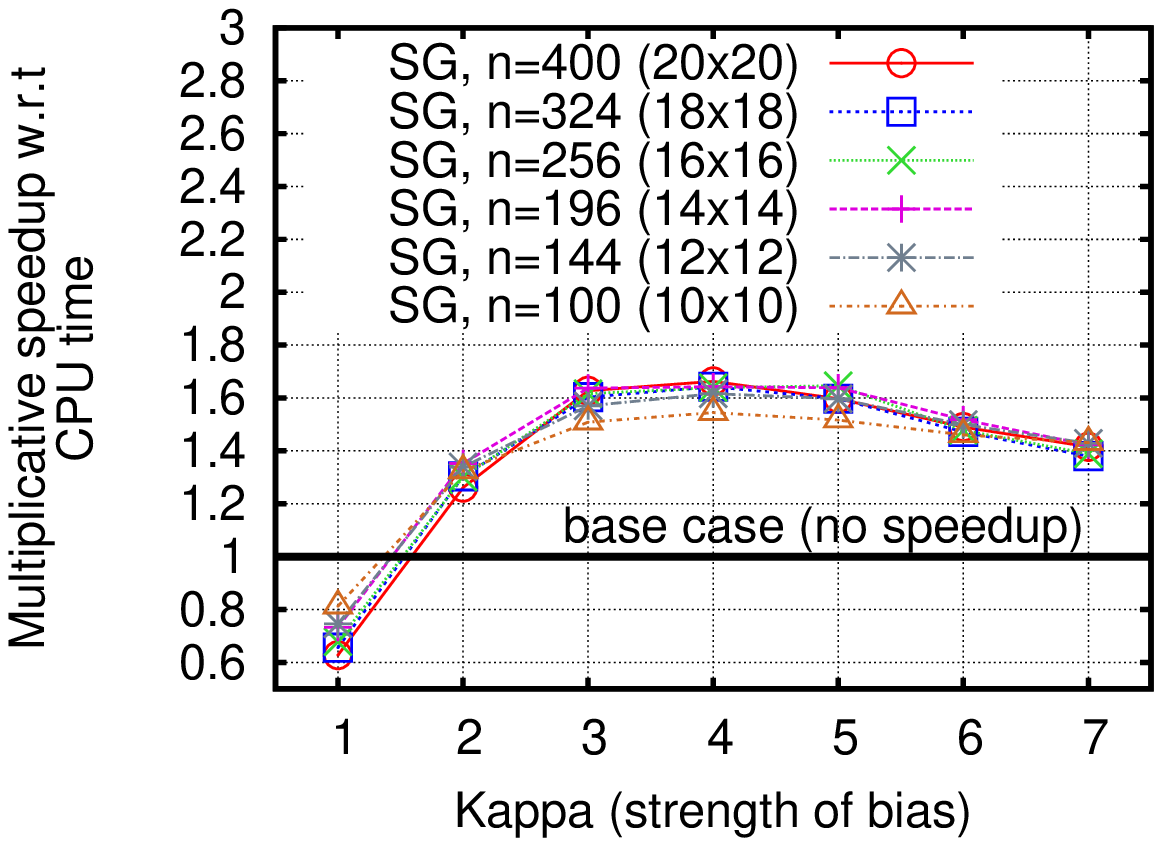,width=0.33\textwidth}
\\
\vspace*{-0.4em}
\epsfig{file=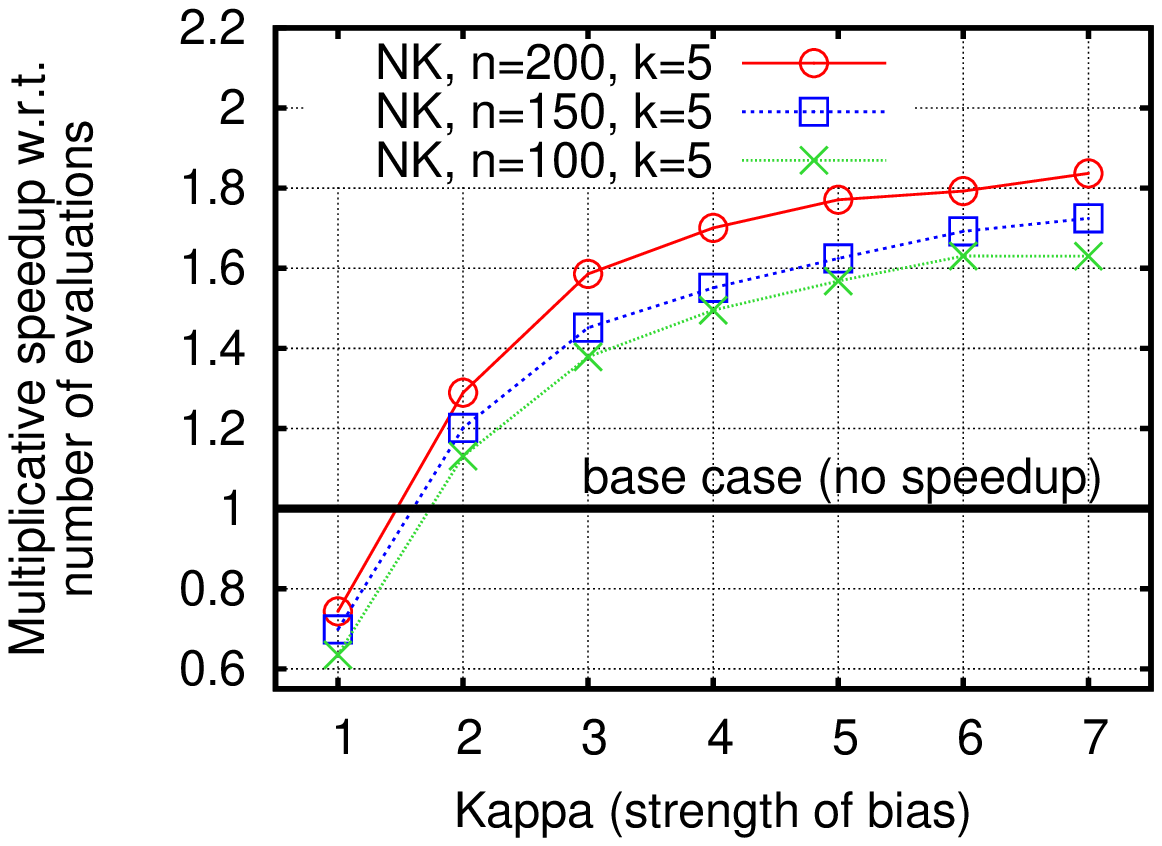,width=0.33\textwidth}\hspace*{5em}
\epsfig{file=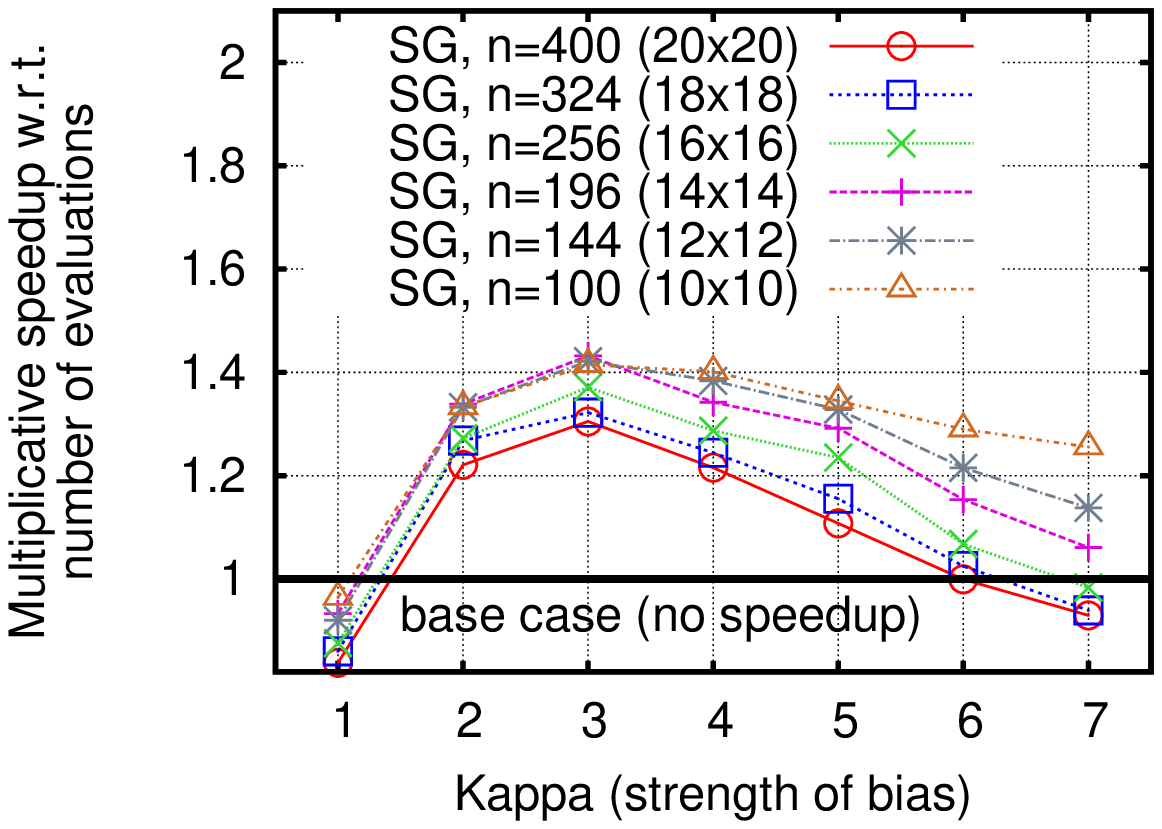,width=0.33\textwidth}
\\
\vspace*{-0.4em}
\epsfig{file=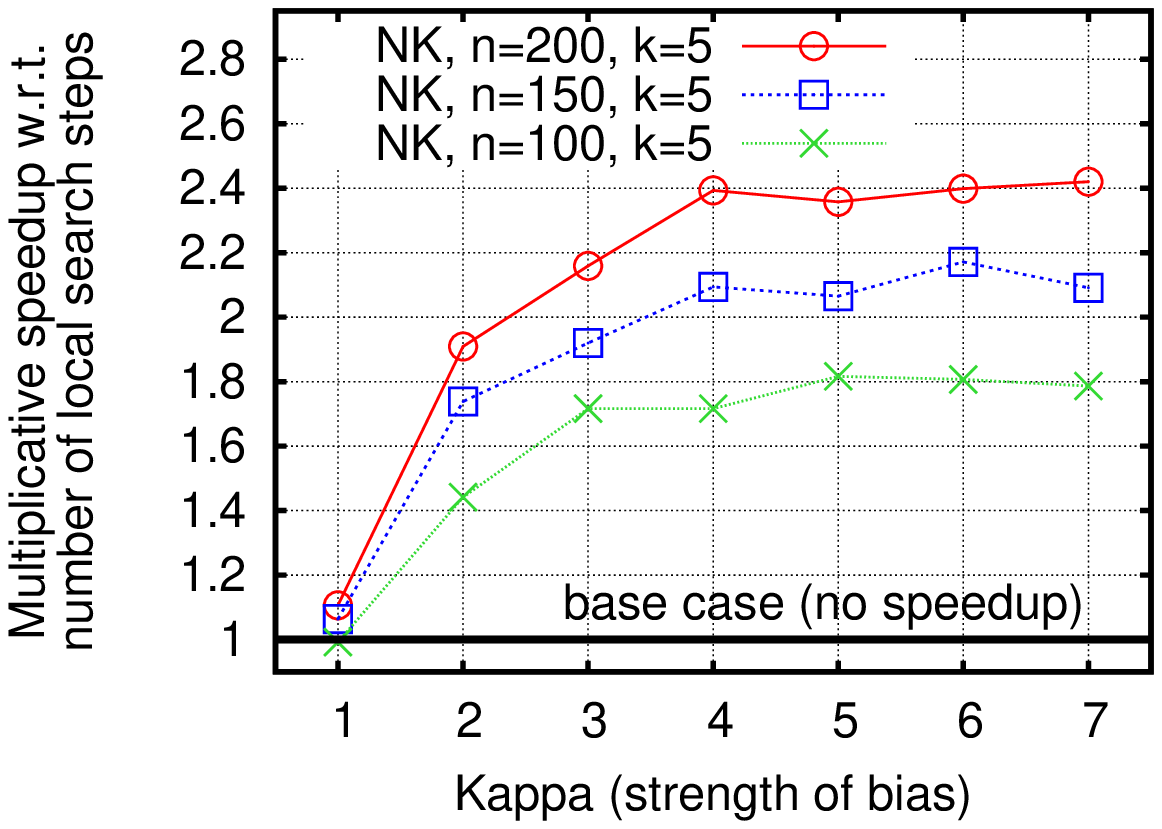,width=0.33\textwidth}\hspace*{5em}
\epsfig{file=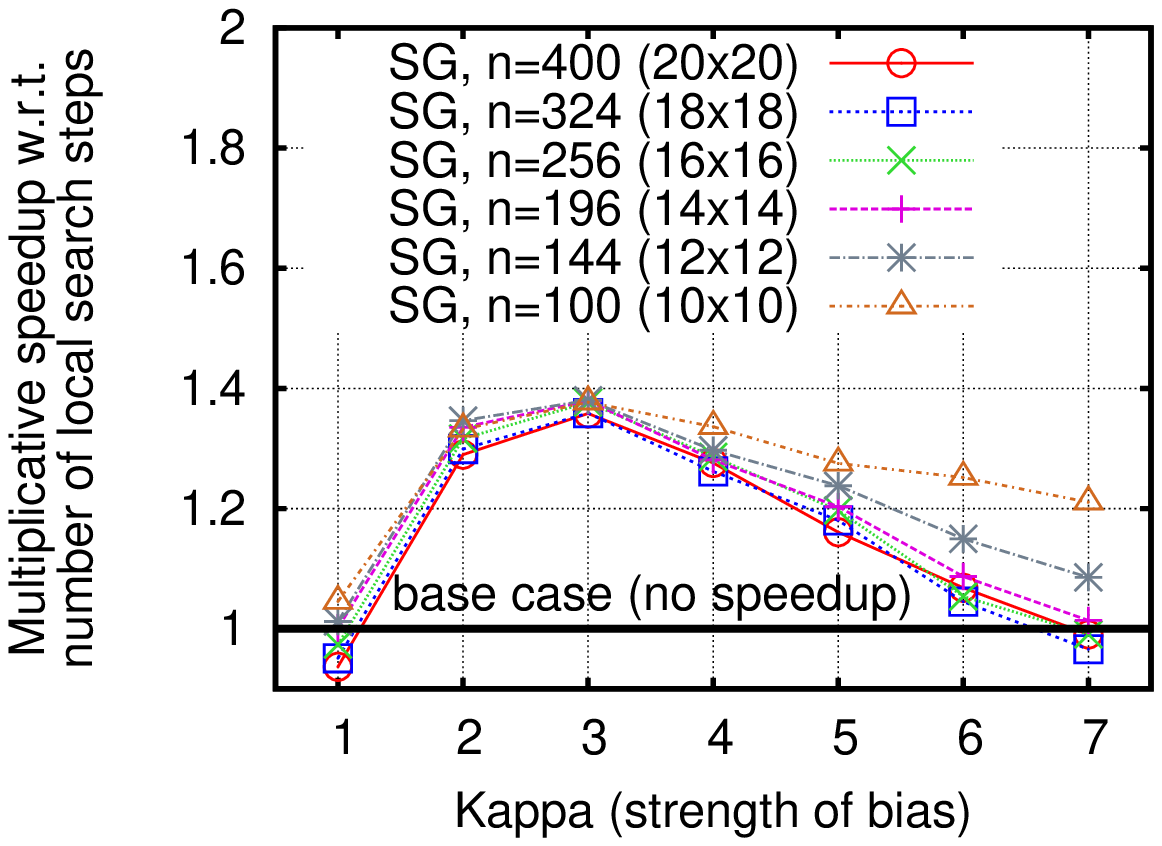,width=0.33\textwidth}
\vspace*{-0.85em}
\subfigure[Effects of $\kappa$ on the reduction of the execution (CPU) time, the number of evaluations, the number of steps of local search, and the population size.]{
\label{fig-speedups1a}
\vspace*{-0.7em}
\epsfig{file=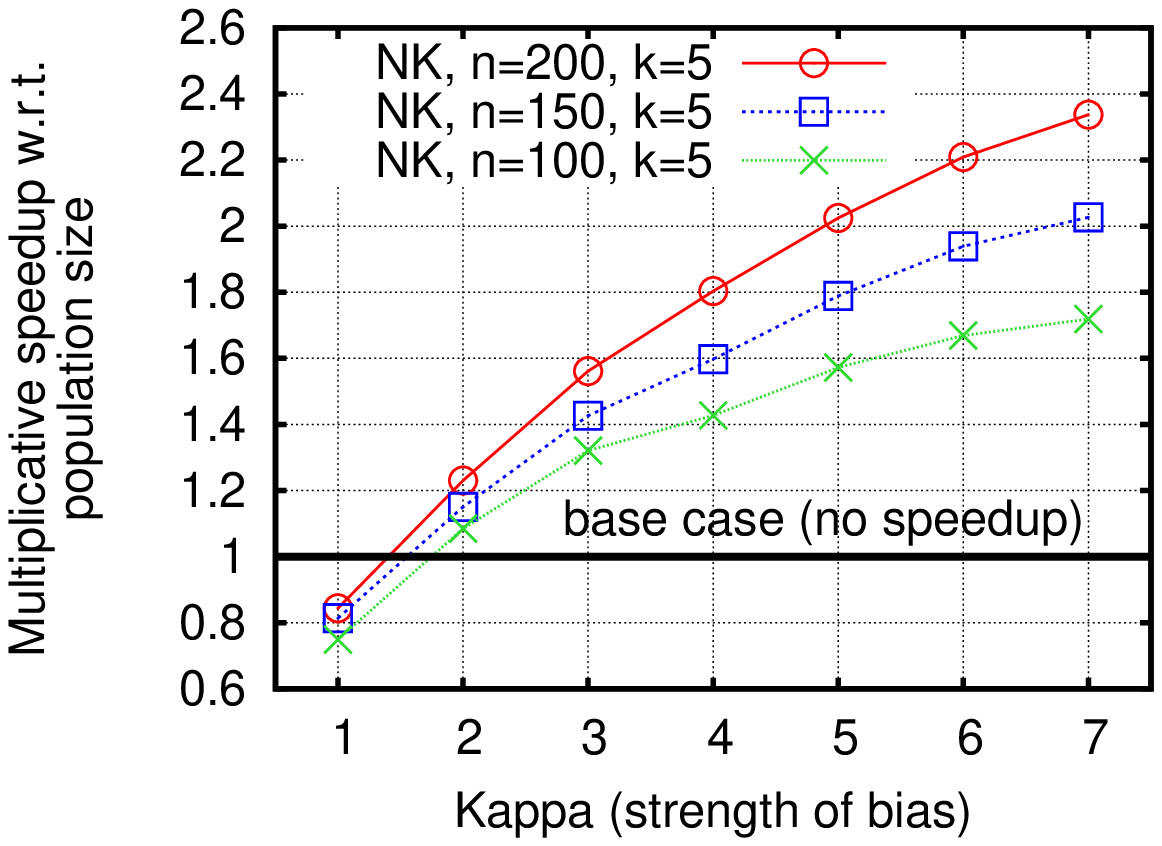,width=0.33\textwidth}\hspace*{5em}
\epsfig{file=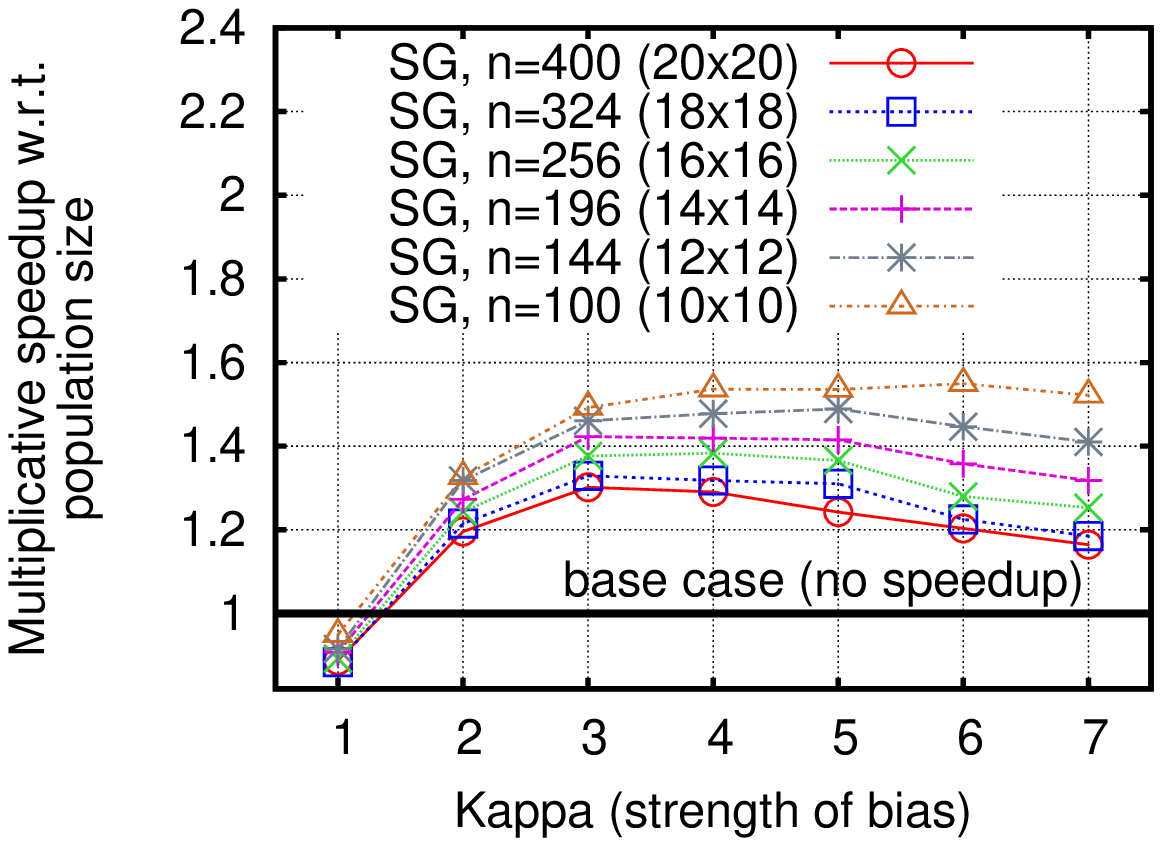,width=0.33\textwidth}
}\\

\vspace*{-0.05em}

\subfigure[Effects of problem size on execution-time speedups.]{
\label{fig-speedups1b}
\epsfig{file=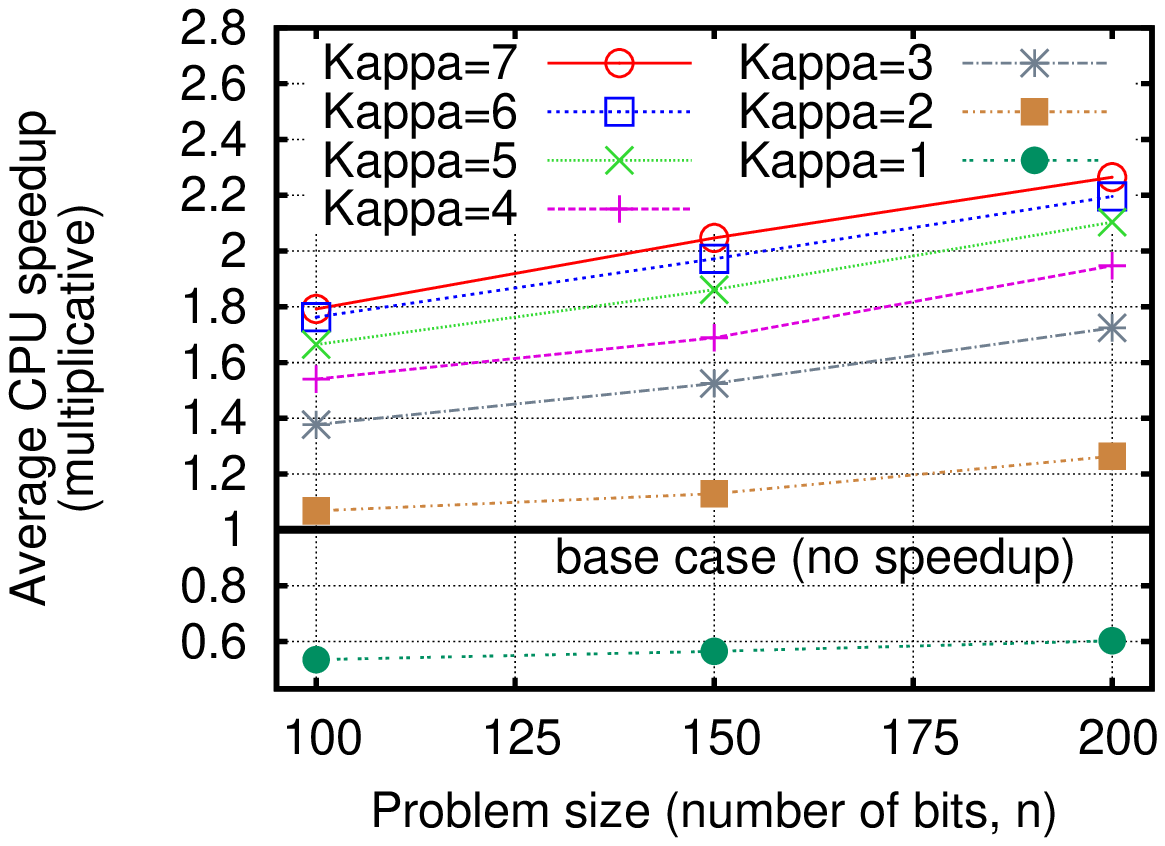,width=0.33\textwidth}
\hspace*{5em}
\epsfig{file=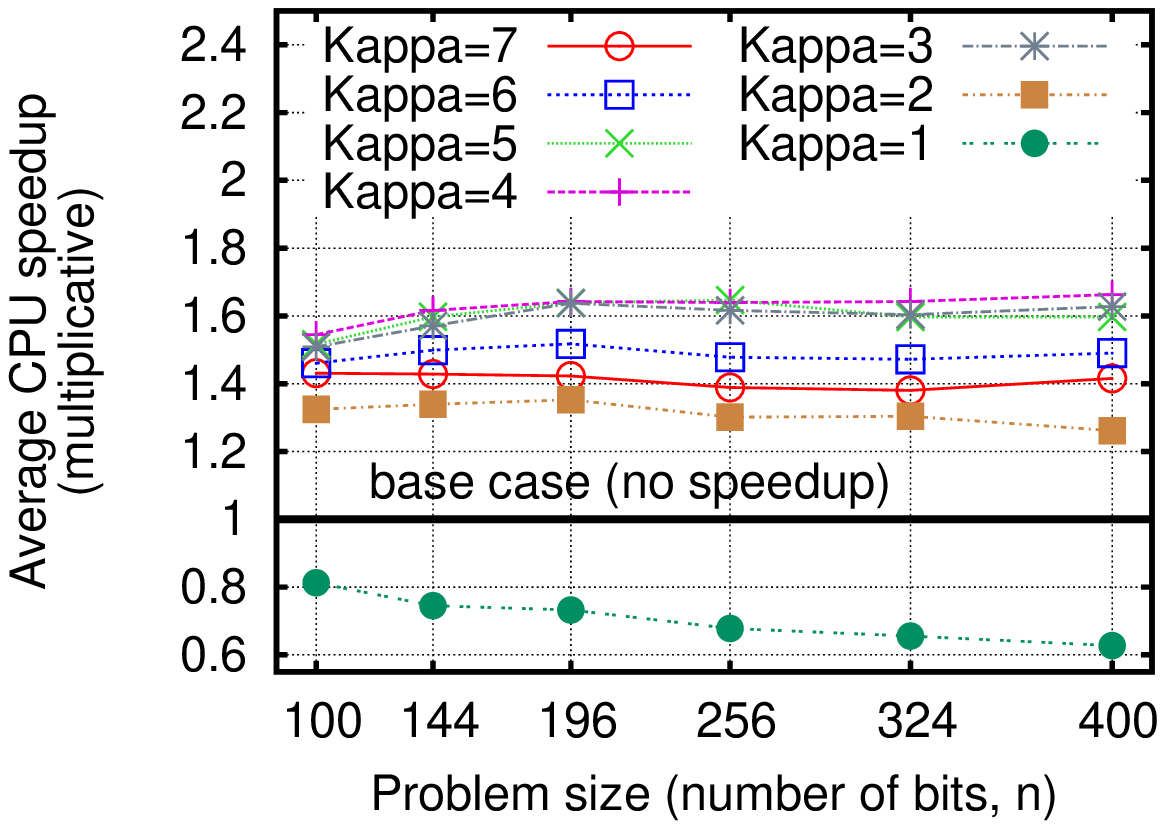,width=0.33\textwidth}
}
\vspace*{-0.85em}
\caption{\small Effects of the distance-based bias based on models from hBOA runs on other problem instances measured by multiplicative speedups. The speedup for a particular statistic is the ratio of the value of the statistic without the bias and its value with the bias. 
Thus, the greater the speedup, the better the effects of the bias. For example, if CPU speedup is 2, hBOA with the distance-based bias is twice as fast as hBOA without it in terms of the total execution time. The results for NK landscapes are shown on the left, the results for 2D spin glasses are shown on the right.}
\label{fig-speedups}
\end{center}
\end{figure*}



\section{Summary and Conclusions}
\label{section-conclusions}
This paper introduced a practical approach to incorporating bias in estimation of distribution algorithms (EDAs) based on models built in previous EDA runs on problems of similar type. The approach was demonstrated on the hierarchical Bayesian optimization algorithm (hBOA) and additively decomposable functions, although the framework can be applied also to other EDAs based on graphical models and other problem types. For example, it should be straightforward to adapt this framework to the extended compact genetic algorithm~\cite{Harik:99} or classes of facility location problems. The key idea of the proposed approach was to define a distance metric that corresponds to the likelihood of dependencies between variables, and to use the statistics on dependencies at various distances in previous hBOA runs as the basis for introducing bias in future hBOA runs. The bias was introduced using prior probabilities of Bayesian network structures. The models were thus learned using a combination of  the selected population of candidate solutions and the prior knowledge extracted from previous hBOA models. The strength of the bias can be tuned with a user-defined parameter $\kappa>0$. 
The proposed approach was tested on two challenging additively decomposable functions, the NK landscapes with nearest-neighbor interactions and the two-dimensional Ising spin glass. 
The results on 9,000 unique problem instances from the two problem classes provided empirical evidence that the proposed approach provides substantial speedups across a variety of settings. Specifically, speedups of over $2.26$ were achieved for NK landscapes, and speedups of over $1.66$ were achieved for spin glasses. Furthermore, the speedups for NK landscapes grew with problem size and the parameter $\kappa$, and it can thus be expected that higher speedups can be achieved in practice. Preliminary experiments on other problem classes, including MAXSAT and minimum vertex cover, indicate that the approach provides substantial benefits also in other important classes of additively decomposable functions. 

The results thus reaffirm that one of the key advantages of EDAs is that EDAs provide practitioners with a rigorous framework for incorporating prior knowledge and for automated learning from solving instances of similar type  so that future problem instances can be solved with increased speed, accuracy, and reliability. EDAs thus not only allow practitioners to scalably solve problems with high levels of epistasis (variable interactions), but they also allow effective inductive transfer (transfer learning) in optimization. 

In future work, the approach should be tested on other additively decomposable problems. Experiments should also be done to confirm the hypothesis that models obtained on problems of one size can be used to bias model building on problems of another size. Furthermore, the approach should be adapted to other model-directed optimization techniques, including other EDAs and genetic algorithms with linkage learning. The approach should also be modified to introduce bias on problems that cannot be formulated using an additive decomposition in a straightforward manner. Finally, it is important to study the limitations of the proposed approach, and create theoretical models to automatically tune the strength of the bias and predict expected speedups. 



\section*{Acknowledgments}
This project was sponsored by the National Science Foundation under grants ECS-0547013 and IIS-1115352, and by the University of Missouri in St. Louis through the High Performance Computing Collaboratory sponsored by Information Technology Services. Most experiments were performed on the Beowulf cluster maintained by ITS at the University of Missouri in St. Louis and the HPC resources at the University of Missouri Bioinformatics Consortium. hBOA was developed at the Illinois Genetics Algorithm Laboratory at the University of Illinois at Urbana-Champaign. Any opinions, findings, and conclusions or recommendations expressed in this material are those of the authors and do not necessarily reflect the views of the National Science Foundation. 


\bibliographystyle{apa-uiuc}
\bibliography{mybib}

\end{sloppy}

\end{document}